# A Machine-learning Framework for Automatic Reference-free Quality Assessment in MRI

T. Küstner[a,c,*], S. Gatidis[b], A. Liebgott[a,b], M. Schwartz[a,c], L. Mauch[a], P. Martirosian[c], H. Schmidt[b], NF. Schwenzer[b], K. Nikolaou[b], F. Bamberg[b], B. Yang[a], F. Schick[c]

[a]*Institute of Signal Processing and System Theory, University of Stuttgart, Stuttgart, Germany*
[b]*Department of Radiology, University of Tübingen, Tübingen, Germany*
[c]*Section on Experimental Radiology, University of Tübingen, Germany*

## Abstract

Magnetic resonance (MR) imaging offers a wide variety of imaging techniques. A large amount of data is created per examination which needs to be checked for sufficient quality in order to derive a meaningful diagnosis. This is a manual process and therefore time- and cost-intensive. Any imaging artifacts originating from scanner hardware, signal processing or induced by the patient may reduce the image quality and complicate the diagnosis or any image post-processing. Therefore, the assessment or the ensurance of sufficient image quality in an automated manner is of high interest. Usually no reference image is available or difficult to define. Therefore, classical reference-based approaches are not applicable. Model observers mimicking the human observers (HO) can assist in this task. Thus, we propose a new machine-learning-based reference-free MR image quality assessment framework which is trained on HO-derived labels to assess MR image quality immediately after each acquisition. We include the concept of active learning and present an efficient blinded reading platform to reduce the effort in the HO labeling procedure. Derived image features and the applied classifiers (support-vector-machine, deep neural network) are investigated for a cohort of 250 patients. The MR image quality assessment framework can achieve a high test accuracy of 93.7% for estimating quality classes on a 5-point Likert-scale. The proposed MR image quality assessment framework is able to provide an accurate and efficient quality estimation which can be used as a prospective quality assurance including automatic acquisition adaptation or guided MR scanner operation, and/or as a retrospective quality assessment including support of diagnostic decisions or quality control in cohort studies.

## 1. Introduction

Magnetic Resonance Imaging (MRI) is an imaging modality providing a variety of contrast mechanisms by which one can visualize both anatomical structures and physiological functions inside the human body. The immense and flexible MR sequence and reconstruction parametrization makes it on the one hand flexibly tunable to specific needs and applications but demands a profound technical knowledge on the other hand. Besides its various advantages, MR acquisition takes rather long and images are often prone to artifacts due to hardware imperfections like magnetic field inhomogeneities, applied signal processing like image aliasing or patient compliance like bulk or respiratory motion. Hence for a good image quality, a careful MR sequence and reconstruction selection and parametrization have to be performed, and optimal acquisition conditions have to be ensured to reduce artifacts. Since not all objectives can be fulfilled simultaneously, one aims for the best achievable image quality with respect to a certain application by compromising a tradeoff. Due to the large MR sequence and reconstruction variety, nowadays enormous amounts of data are created per patient. So far image quality evaluation is a manual process which mainly depends on human observers (HO) like trained physicians or experienced radiologists. The considered quality criteria need to be clarified and ensured first according to specific conditions or diagnostic questions. Thus, this process can be very time consuming and cost-intensive.

---

*Corresponding author
  *Email address:* thomas.kuestner@iss.uni-stuttgart.de (T. Küstner)

An objective image quality evaluation is often very demanding, especially when developing new technologies or methods for which no reference or gold-standards are available yet. A prolonged scan time or non-compliant patients make the acquisition of reference images very difficult.

There is a variety of applications in which an automatic medical image quality assessment could be of potential interest. These applications can be grouped into i) prospective quality assurance and ii) retrospective quality assessment/control. In a prospective manner, nowadays nearly none or just little feedback of the acquired image quality is provided back to the MR scanner or its operating technician if the current acquisition falls below a defined quality threshold, e.g. artificial burden in the targeted imaging region. A direct feedback to rerun the scan with possible reparametrization while the patient is still in the scanner will help to ensure a high image quality with reduced costs. A retrospective quality control can support a guided or ordered image reading of the HO increasing the patient comfort and throughput. Moreover, for large cohort studies [1, 2] or in quantitative measurements a reliable and repeatable image quality has to be guaranteed. Any image post-processing or analysis can only be correctly applied to images of sufficient quality.

For a general image quality assessment, it is important to evaluate the images based on the image quality depending on the underlying question or application of interest such as e.g. diagnostic usability, artifact burden, quantitative reliability and so on. State-of-the-art methods differ in terms of i) being task-specific or minimal-generic, ii) based on a hand-crafted model or a machine-learning approach with iii) available reference data or in a reference-free setting. In early works, model observers (MO) based on intensity similarity/dissimilarity measures to a reference image were suggested to evaluate the impact of image disturbances [3, 4, 5, 6, 7]. As indicated in [8], exploiting known characteristics of the human visual system (HVS) can improve the image quality assessment performance [9, 10, 11] by better reflecting the human perception, i.e. mimicking the HO. Numerous HVS approaches model the perceived image disturbances such as blurring or noise in order to quantify types and degrees of distortions [12, 13]. These HVS models are based on image-derived statistics [14, 15, 16], a combination of natural-scene statistics and extracted features [17, 12] or holistic models which incorporate several image-derived statistics [18]. In order to reflect the HVS and mimicking the HO, learning from labeled HO data is crucial. In the work of [19] a semi-supervised learning was proposed for non-redundant sample selection in the training procedure. In the work of Lorente et al. [20], an active learning was applied to a task-based quality assessment with a relevance vector machine in a four-dimensional feature space. In simpler disturbance scenarios, [21] presented a method to omit the HO labeling completely by an automatic learning from overlapping image patches.

In a medical context, most works focused on automatic lesion detection as a binary classification problem [22, 23]. A channelized hotelling observer (CHO) [24], a relevance vector machine [25] or a machine-learning approach [26] were used to detect cardiac defects in single-photon emission computed tomography (SPECT) images. The CHO was further studied in [27] to determine a quantitative image metric based on a perceptual difference model. The work in [28] proposed a univariate signal analysis of the image background for brain images. The quality estimation task for neuroimaging databases was extended in [29] by extracting features to train a support vector machine (SVM) for predicting artifact burden.

The aim of this study is to provide a medical MR image quality assessment framework which reflects the HVS. It can be applied for quality assessment of MR images in large cohort studies [2] to ensure a correct processing of any subsequent method. More complicated image distortions, like motion or subsampling-related artifacts degrade the perceivable image quality and cannot be captured easily by any of the previous methods [30]. Since the image quality assessment shall reflect the HVS, the model is trained on HO-derived labels. The spent HO labeling effort (time and costs) is a crucial factor in the whole framework. In a first study [31], we investigated the applicability of a machine-learning approach to predict HO labeling scores of MR images on a discrete Likert-scale. This study aims to derive MR image quality from images with complex distortions (motion and subsampling artifacts) of arbitrary input (body region, imaging sequence) with and without acquisition acceleration (Parallel Imaging; PI and Compressed Sensing; CS) in a blind/reference-free setting. Instead of a binary decision (i.e. affected or not affected by artifacts), we provide a quality output on $K$ Likert-scale classes [32] enabling a more fine-grained distinction. The framework is trained on limited training data to reduce the effort in the HO labeling step. Opposed to the previous works [20, 28, 29], the classification is conducted in a high dimensional feature space in order to model complex image distortions in arbitrary body regions and from various acquisition sequences. The required feature space dimensionality to produce a reliable generalization is also investigated.

In this work we will highlight two aspects to create a generic and robust MR image quality assessment framework. The first contribution is the extension and successful conjunction of the individual components which are required for a cohort's image quality assessment. Several distinct image features are extracted from the input data to reflect different characteristcs [33]. In the classification, a one-vs.-one multi-class soft-margin SVM is compared to a deep neural network (DNN) [34]. In order to reduce the amount of needed labeled training data, the MR image quality assessment framework possesses an efficient active learning procedure [35, 36, 37] which is extended for 3D images and equipped with a probability-based uncertainty selection.

As a second and main contribution we investigate the effectiveness to perform the desired quality assessment task. In this context, quality is defined as diagnostic usability which is mainly affected by MRI artifacts. The framework predicts scores which are compared to HO-derived labeling scores. In a cohort study of 508 patients, the suitable feature combinations in interaction with the employed classifiers and the possible achievable training data reduction by active learning is investigated. The derived results allow the formulation of a reliable and robust MR image quality assessment which may be used in the future to prospectively assure image quality directly on the MR scanner or to analyze obtained image quality retrospectively on a daily basis or in cohort studies.

## 2. Material and Methods

A block diagram of the proposed framework is shown in Fig. 1. The architecture is chosen to ensure a generic framework which can be used for any image quality assessment application. In this work however we investigate solely the scoring of image quality in terms of diagnostic usability. The illustrated blocks will be described below.

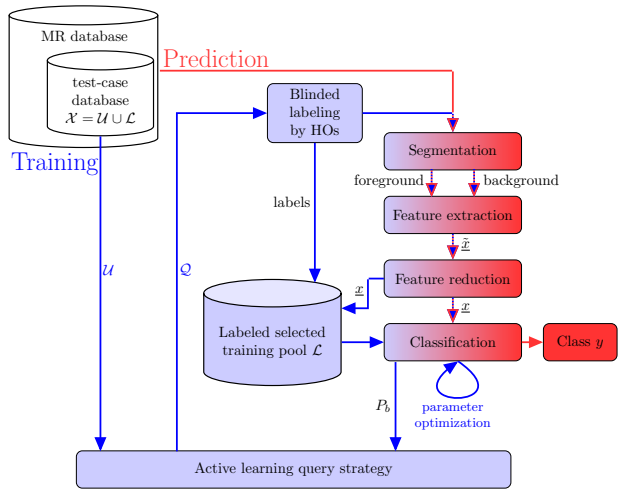

Figure 1: Image quality assessment framework for training (blue) and prediction (red). The test-case database $\mathcal{X}$ consists of the dynamic unlabeled datasets $\mathcal{U}$ and labeled datasets $\mathcal{L}$. In the training, in each active learning iteration a set of query images $\mathcal{Q}$ is labeled and inserted into the labeled training set $\mathcal{L}$ after feature extraction and reduction of the feature vector $\underline{\tilde{x}} \in \mathbb{R}^F$ towards $\underline{x} \in \mathbb{R}^R, R < F$. In the classification, an unknown image is described via $\underline{x}$ and classified into a quality class $y \in [1, K]$ depending on the trained question/application.

*2.1. MR database*

The MR database should reflect the to be learned quality criteria of the MR images. It should thus contain 3D or 2D multislice sequences from a static or dynamic imaging for different body regions, artifact levels, acquisition and reconstruction strategies. Hence, the collection of the database and the selection of MR samples from the database for training is an important step. Moreover, a large database is needed to provide sufficient diversity and generalizabilty which is especially important for a DNN-based classification. This results in a great effort in the labeling step. We will address this issue by using active learning [35, 36] which tries to select the most meaningful samples from the database, see 2.7.

From the MR database of all acquired datasets, a selected subset of samples is chosen for the training to uniformly represent different criteria reflecting the underlying question/application, e.g. body regions and acquisition strategies. This subset forms the so called test-case database $\mathcal{X} = \mathcal{U} \cup \mathcal{L}$ consisting of a dynamic (changing in size) subset $\mathcal{U}$ of unlabeled datasets and a dynamic subset $\mathcal{L}$ of labeled training datasets. In the beginning no dataset is yet labeled and hence all data belongs to the unlabeled set $\mathcal{U} = \mathcal{X}, \mathcal{L} = \emptyset$. $\mathcal{L}$ increases by the same amount of datasets as $\mathcal{U}$ decreases during labeling and active learning. At some point all datasets are labeled ($\mathcal{L} = \mathcal{X}$) and there can be no further classification improvement by the active learning dataset selection ($\mathcal{U} = \emptyset$). Note that a dataset or sample is an element of the database and the smallest entity whose image quality has to be scored. In this work, a dataset is chosen to be the complete MR image series of a 3D or 2D multislice image acquisition which is sufficient for the task of overall observable image quality. Although some artifacts may only affect part of the image volume, these local artifacts still depreciate the overall perceivable image quality and hence the HOs tend to score image quality on a global view. This carefully selected test-case database $\mathcal{X}$ ensures sufficient diversity and resemblance to the underlying question/application. During the prediction phase, an unknown dataset from the MR database is scored by the trained framework with respect to the underlying question/application.

*2.2. Labeling*

Since the MR image quality assessment framework relies on supervised learning to mimic the HVS, the unlabeled datasets/image series in $\mathcal{U}$ are presented to an HO for a blinded scoring according to an underlying question/application as specified in the test-case. The labeled image-series are then added to the training pool $\mathcal{L}$.

With the applied active learning concept (see 2.7) only small dataset packages are labeled per query. To streamline the blinded labeling procedure, an HTML-based website is developed which can be easily accessed from every computer within the hospital [37]. HOs register to the website and provide some background information (e.g. field and years of experience). Depending on the determined study prerequisites, e.g. specific required expertises, eligible HOs receive an email invitation to participate. Before labeling, the HOs are presented with a detailed instruction page of the test-case composition and interrogation to minimize the inter-reader ambiguity. This instruction page can be viewed anytime for guidance. The labeling can be stopped at any moment and resumed at a later point. Previously labeled datasets of the same HO can be examined to guarantee intra-reader consistency.

The HO labels each dataset by assigning it to one of the $K$ predefined quality classes. The meaning of the classes is determined by the test-case composition and communicated to the HO.

*2.3. Foreground/background segmentation*

While in MRI the foreground holds the principal image content, the background reveals information about artifacts (infolding, aliasing, motion ghosting, ...) and noise level. Thus, each image is first divided into a foreground and a background using an active-contour Chan-Vese segmentation [38] with a rounded rectangular mask as initialization. Distinct information are extracted from these two regions in the subsequent feature extraction which help to further score datasets by a classification.

*2.4. Feature extraction*

An image can be described by many different attributes such as smoothness, coarseness, regularity, brightness, homogeneity and so on. They lead to a deeper understanding of the actual image and describe it by other attributes and characteristics to enable the differentiation and classification of images in a large dimensional space. The task of feature extraction is thus to define and calculate highly discriminative features from the MR image which help to maximize the inter-class and minimize the intra-class distance. The selected features directly influence the performance of the image quality assessment and have thus to be determined and tuned carefully according to the underlying question/application at hand.

For sufficient diversity and specificity to discriminate different classes, features used in the classification are determined from a large pool of possible features forming the original feature vector $\tilde{\underline{x}} \in \mathbb{R}^F$. An optimal feature selection will be investigated in this work. For a better understanding, they are divided into different groups as illustrated in Fig. 2. For each group the main methods for calculating features from an image are depicted. The main distinction is between global and local features reflecting the extracted image information content.

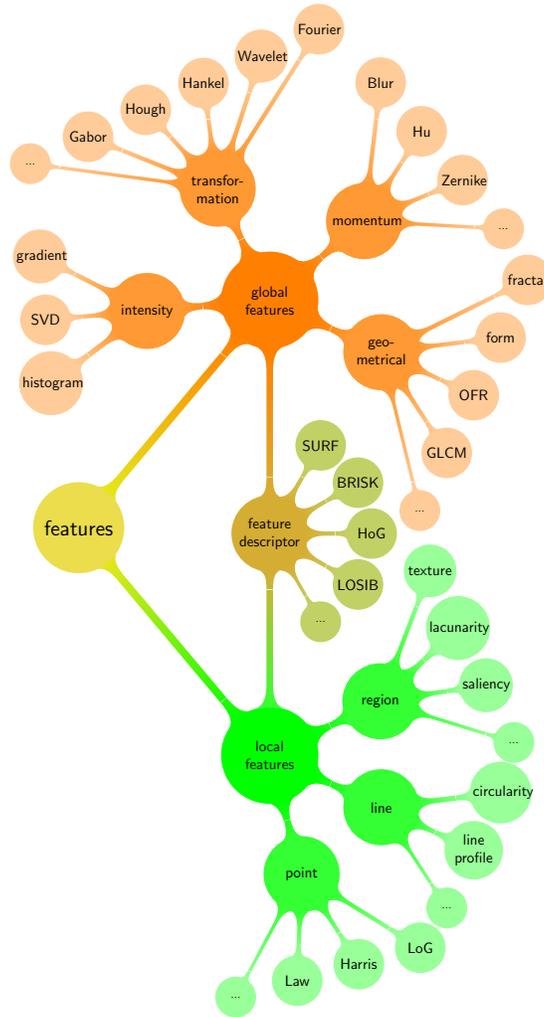

Figure 2: Overview and connection of different groups of features.

- Global features act on the complete image content independent of any local neighbourhood criteria. They provide a coarse description of the image and can be further grouped into intensity-based, transformation, momentum and geometrical features.

- Local features are calculated directly from image segments or from implicitly derived spatial informa-

tion and contain local information like neighbourhood. Local features describe the image in a more detailed view. The corresponding subgroups reflect either point, line or region information.

- The group of feature descriptors summarize powerful features to describe important key points and/or regions on a global and/or local scale. Feature descriptors provide e.g. semantic characterization of objects.

An overview of all derived features and the used methods are shown in Table 1. A detailed description for each of the features can be found in the referenced literature. Features marked with (∗) are self-defined and mainly focus on extracting energy and entropy information from specific transformation ranges which reflect certain structures such as circles, lines or homogeneous regions. Furthermore, these features can be extracted from different orthogonal image directions or with different algorithmic parametrizations, e.g. distinct wavelets. The resulting feature vector of length $F$ is denoted by $\underline{\tilde{x}} \in \mathbb{R}^F$.

*2.5. Feature reduction*

In order to avoid overfitting and redundancy and to reduce the computational complexity, the dimension of the feature space is reduced by a principal component analysis (PCA) [69]. Each original feature vector $\underline{\tilde{x}}_n$ of sample $n$ is projected onto an $R$-dimensional subspace $\mathbb{R}^R$ of space $\mathbb{R}^F$ spanned by the $R$ dominant eigenvectors of the sample covariance matrix of the zero-mean and unit variance scaled feature vectors $\underline{\tilde{x}}_n$. The resulting PCA transformed feature vector is denoted by $\underline{x}_n \in \mathbb{R}^R$ and serves as input for the following classification.

*2.6. Classification*

In a supervised fashion, during training phase a classifier model is trained according to the labeled feature vectors $\underline{x}_n, n \in [1, |\mathcal{L}|]$ with $|\mathcal{L}|$ being the amount of all available training datasets. Two classifiers are investigated, a multi-class soft-margin SVM and a DNN. Each feature vector is classified into one of the $K$ discrete quality classes $y \in [1, K]$.
After training, in the prediction phase the learned classifier model is applied to an unlabeled dataset to predict the most likely quality class $\hat{y}$.

*2.6.1. Support vector machine*

The multi-class SVM uses an one-against-one approach to make a majority-based combination of the results of $K(K-1)/2$ binary classifications [70]. Each binary classification uses a soft-margin SVM with the penalty regularization constant $C$. In order to perform a non-linear feature mapping, a radial basis function (RBF) kernel $k(\underline{x}_n, \underline{x}_m) = \exp(-\gamma \|\underline{x}_n - \underline{x}_m\|_2^2)$ with the kernel parameter $\gamma$ is used.

*2.6.2. Deep Neural Network*

The DNN is a feedforward multilayer network with $M$ layers ($M-2$ hidden layers) and $N_n^{(m)}$ nodes in each layer. Each node on one layer is fully connected to all nodes of the previous layer. A non-linear and continuously differentiable exponential linear unit activation function is used to map the inputs to the outputs.

For the final output layer, a softmax function is used to estimate the class probabilities from which the most likely class determined the label $\hat{y}$. The categorial cross-entropy cost function is minimized via an adaptive moment estimation (ADAM) [71] with associated ADAM parameters $\beta_1, \beta_2$ and $\epsilon$ under an $\ell_2$ regularization. In addition a dropout ratio per layer $m$ is used. The weights and bias values are initialized by a Gaussian distributed randomization with input size scaling [72] to enable a fast convergence.

Table 1: Number of generated image features (indicated by #) and the used methods in the feature extraction step categorized into groups following Fig. 2. (∗) marks self-defined features.

| | method | ref. | # | | | method | ref. | # |
|---|---|---|---|---|---|---|---|---|
| **intensity** | gradient | [39] | 24 | | **feature descriptor** | Speeded Up Robust Features (SURF) | [40] | 152 |
| | singular value decomposition (SVD) | ∗ | 660 | | | Binary Robust invariant scalable points (BRISK) | [41] | 13 |
| | histogram | [42],∗ | 10 | | | Histogram of oriented gradients (HoG) | [43] | 2304 |
| **global features / transformation** | Fourier | ∗ | 300 | | | Local Oriented Statistic Information Booster (LOSIB) | [44] | 34 |
| | Discrete Cosine | ∗ | 3447 | | **region** | Region Covariance descriptors (RCovDs) | [45] | 7 |
| | Wavelet | ∗ | 4623 | | | edge-based region (EBR) / intensity extrema-based region (IBR) | [46] | 1387 |
| | Hankel | ∗ | 75 | | | local binary pattern (LBP) | [47] | 1024 |
| | Distance | ∗ | 52 | | | Maximal Stable Extremal region (MSER) | [48] | 15 |
| | Hough | ∗ | 376 | | | gray-level co-occurence matrix (GLCM) | [42] | 651 |
| | Top-Hat | ∗ | 60 | | | salient region detector | [49] | 8 |
| | Gabor | [50] | 1080 | | | lacunarity | [51] | 66 |
| | Walsh-Hadamard | ∗ | 5 | | | connectivity | [52] | 155 |
| | Hilbert | ∗ | 5 | | | Quad-tree decomposition | [53] | 5 |
| | Chirp-Z | ∗ | 58 | | **line** | line profile | ∗ | 122 |
| | Radon | ∗ | 5 | | | line & circle detection | [54] | 9 |
| | Skeletonization | [55] | 8 | | | edges (Sobel, Canny, Prewitt) | ∗ | 15 |
| **moment** | Zernike | [56] | 80 | | | Ant-swarm edges | [57] | 5 |
| | Hu | [58] | 8 | | | Harris detector | [59] | 11 |
| | Blur and Affine | [60] | 6 | | | Law | [61] | 50 |
| **geometrical** | gray-level co-occurence matrix (GLCM) | [42] | 21 | | **point** | Laplacian of Gaussian (LoG) | [62] | 260 |
| | run length | [63] | 44 | | | Smallest univalue segment assimilating nucleus (SUSAN) | [64] | 4 |
| | optical font feature (OFR) | [65] | 22 | | | Gilles | [66] | 27 |
| | fractal dimension | [67] | 27 | | | | | |
| | form factor | [68] | 66 | | | | | |

$\sum$ **17386**

### 2.7. Active Learning

Instead of labeling all datasets which can be very time-consuming and cost-intensive, active learning employs query strategies to find those datasets which are expected to have the most positive influence on the performance of the classifier. Thus, active learning is utilized to select only the most discriminative datasets which provide non-redundant information to the classification. The previously published active learning method [35, 36, 37] is extended to the 3D case and utilizes a probability-based uncertainty.

From the pool of unlabeled datasets $\mathcal{U}$, $N_\text{I}$ datasets (image-series) are randomly selected for an initial training and inserted into the training pool $\mathcal{L}$ after labeling, feature extraction and reduction. The goal of active learning is to keep $\mathcal{L}$ as small as possible by selecting the most informative datasets for training. This avoids redundancy while still providing a high classification accuracy. The selection of new image-series for active learning, also called query strategy, is based on the idea of uncertainty sampling [73], i.e. selecting image-series for which the classifier is most uncertain about. At each active learning iteration, image-series in $\mathcal{U}$ are selected to form the query set $\mathcal{Q}$. The uncertainty selection is based on the probabilities [74] of the most probable classes and can be easily extended to the multi-class case by pairwise coupling [75].

Let $\underline{x}$ be the feature vector after segmentation, feature extraction and reduction (see 2.3 to 2.5) of a 2D image from a dataset $I$ in $\mathcal{U}$. Each feature vector $\underline{x}$ is classified by the classifier of the previous active learning iteration. Let $P_{b1}(\underline{x})$ and $P_{b2}(\underline{x})$ be the probability of the most and second most probable class, respectively. The MR image represented by $\underline{x}$ is defined to be uncertain with respect to the current classifier if $P_{b1}(\underline{x}) - P_{b2}(\underline{x})$ is small. By selecting the slices belonging to the most beneficial dataset $I_n$, one can gain a measure of uncertainty for the image selection from the unlabeled set $\mathcal{U} = \mathcal{X} \setminus \mathcal{L}$ for the query set

$$\mathcal{Q} = \bigcup_{n=1}^{|\mathcal{Q}|} \{I_n \in \mathcal{U} | \arg\min_n \sum_{\underline{x}_m \in I_n} \min_m (P_{b1}(\underline{x}_m) - P_{b2}(\underline{x}_m))\} \quad (1)$$

The HO is presented on the labeling website (see 2.2) the set $\mathcal{Q}$ of uncertain image-series with decreasing probability of classification uncertainty which he must manually label. The $|\mathcal{Q}|$ labeled uncertain datasets are inserted into the training pool $\mathcal{L}$ to update the classifier. Then the next active learning iteration starts until a certain stopping criterion is satisfied.

## 2.8. Experimental Evaluation

In the considered experiment, the MR database contains a total number of 6035 datasets from 508 patients (age 54.5 ± 16.4 years, 219 female). All datasets are 3D and 2D multislice static anatomic MR image-series of the head, thorax, abdomen, pelvis and whole-body (shoulder to upper legs) with different imaging sequences and contrast weights. They are acquired with gradient echo (GRE), spin echo (SE), inversion recovery (IR) and echo-planar imaging (EPI) sequences. These acquisitions use different subsampling strategies and are affected by artifacts with varying degree: rigid head motion, non-rigid respiratory and cardiac motion in the body trunk, flow/pulsation artifacts and aliasing artifacts of CS-accelerated acquisitions. Artifacts are not explicitly enforced and resemble involuntary or unavoidable distortions. Imaging parameters are fairly similar amongst subjects as depicted in Table 2. Image resolution is fixed with varying matrix size ranges to cover the whole body volume. All reconstructions are performed on the scanner by the vendor-provided software or for the CS-accelerated sequences by the use of Gadgetron [76] and CS_MoCo_LAB [77, 78]. Images were acquired on a 3T whole-body PET/MR (Biograph mMR, Siemens Healthineers, Erlangen, Germany) and on a 3T whole-body MR (Skyra, Siemens Healthineers, Erlangen, Germany). The study was approved by the local ethics committee and all patients gave written consent.

Table 2: MR imaging parameters for gradient echo (GRE), spin echo (SE), inversion recovery (IR) and echo-planar imaging (EPI) sequence.

| | | head | thorax | abdomen | pelvis | whole-body |
|---|---|---|---|---|---|---|
| voxel size [mm$^3$] | T1w GRE | 0.8 x 0.8 x 3.0 | 1.2 x 1.2 x 3.0 | 1.2 x 1.2 x 3.0 | 1.3 x 1.3 x 3.0 | 1.2 x 1.2 x 3.0 |
| | T2w SE | 0.7 x 0.7 x 3.0 | 0.7 x 0.7 x 5.0 | 1.4 x 1.4 x 5.0 | 1.2 x 1.2 x 4.9 | 0.8 x 0.8 x 5.0 |
| | T2w IR | 0.7 x 0.7 x 3.1 | 1.0 x 1.0 x 5.0 | 1.0 x 1.0 x 5.0 | 1.0 x 1.0 x 5.0 | 1.0 x 1.0 x 5.0 |
| | T2w EPI | 3.0 x 3.0 x 3.7 | | | | |
| matrix size | T1w GRE | 256 x 174-320 x 40-80 | 384 x 240-276 x 30-88 | 320 x 256-320 x 20-88 | 320 x 270-320 x 40-64 | 320 x 210-240 x 128-192 |
| | T2w SE | 320 x 320 x 33-43 | 320 x 220-320 x 40-64 | 320 x 280-320 x 30-48 | 320 x 250-320 x 21-67 | 320 x 220-320 x 96-208 |
| | T2w IR | 320 x 276-320 x 26-82 | 384 x 256-384 x 34-50 | 384 x 320-384 x 34-50 | 384 x 320-384 x 30-53 | 384 x 240-384 x 41-136 |
| | T2w EPI | 192 x 192 x 40-80 | | | | |
| dimension | T1w GRE | 3D | 3D | 3D | 3D | 3D |
| | T2w SE | 2D | 2D | 2D | 2D | 2D |
| | T2w IR | 2D | 2D | 2D | 2D | 2D |
| | T2w EPI | 2D | | | | |
| orientation | T1w GRE | transverse | transverse | transverse | transverse | transverse |
| | T2w SE | sagittal | transverse | coronal | transverse | transverse |
| | T2w IR | transverse | coronal | coronal | coronal | coronal |
| | T2w EPI | transverse | | | | |
| TE [ms] | T1w GRE | 1,95 | 1,55 | 1,70 | 1,90 | 1,90 |
| | T2w SE | 110.2 | 97.0 | 93.2 | 85.9 | 95.9 |
| | T2w IR | 72.5 | 56.4 | 55.4 | 56.7 | 62.0 |
| | T2w EPI | 31.67 | | | | |
| TR [ms] | T1w GRE | 4.30 | 3.67 | 3.95 | 4.07 | 4.29 |
| | T2w SE | 4234 | 1200 | 1530 | 3870 | 1200 |
| | T2w IR | 7958 | 7212 | 6920 | 6590 | 6450 |
| | T2w EPI | 2030 | | | | |
| flip angle [°] | T1w GRE | 9 | 9 | 12 | 12 | 12 |
| | T2w SE | 140 | 150 | 120 | 150 | 160 |
| | T2w IR | 150 | 130 | 130 | 135 | 135 |
| | T2w EPI | 80 | | | | |
| bandwidth [Hz/px] | T1w GRE | 510 | 510 | 540 | 450 | 450 |
| | T2w SE | 70 | 425 | 510 | 200 | 640 |
| | T2w IR | 220 | 300 | 300 | 300 | 290 |
| | T2w EPI | 1970 | | | | |

*2.8.1. Database and labeling*

From the overall available 6035 datasets in the MR database, 2911 datasets (250 patients, 120 female) were randomly selected for labeling based on the gender, body regions and available imaging sequences to uniformly represent the test-case database $\mathcal{X}$ as shown in Fig. 3. For the selected imaging sequences, 3% of the datasets in the test-case database have no acceleration at all, 76% have a PI acceleration and 21% have a CS acceleration. Datasets from the database $\mathcal{X}$ are randomly divided into non-overlapping training set with $70\% \cdot |\mathcal{X}| = 2038$ datasets, validation set $10\% \cdot |\mathcal{X}| = 291$ and a test set with $20\% \cdot |\mathcal{X}| = 582$ datasets. Datasets from a patient are either sorted into the training/validation set or in the test set, i.e. the performance is measured on unseen patient data. All datasets were scored by five experienced radiologists on a 5-point Likert-scale directly corresponding to $K = 5$ quality classes $y$, ranging from 1 (very good) to 5 (very poor) with respect to the overall diagnostic usability as the underlying question/application. Artifact impact (motion, flow, subsampling-related aliasing) mainly influence this decision. The median of the 5 HO score values was used as label. The inter-HO reliability was examined by a weighted Fleiss' Kappa based on $g$-agreement with $g = 4$ [79], because the unweighted Fleiss' Kappa suffers from the paradox described in [80].

*2.8.2. Active Learning*

To draw conclusions about the minimal possible training pool size $|\mathcal{L}|$ and to retrospectively analyze the active learning impact, the full database $\mathcal{X}$ is labeled first and subsequently split into training, validation and test set. From the training set, datasets are selected based on the active learning query strategy and added to the training pool $\mathcal{L}$. The minimal possible training pool size $|\mathcal{L}| = N_\mathrm{I} + N_q \cdot |\mathcal{Q}|$ which is composed of the initial training size $N_\mathrm{I}$ and $N_q$ queries of the query size $|\mathcal{Q}|$ was investigated. The test set accuracy

$$\mathrm{ACC} = \frac{|\{\hat{y}_v = y_v\}|}{|\mathcal{V}|} \qquad (2)$$

of the classifier (SVM or DNN) measures the agreement of the estimated Likert-scale class label $\hat{y}_v$ with the true label $y_v$ in the test set $\mathcal{V}$. The minimum training pool size is defined as that value of $|\mathcal{L}|$ for which ACC $> 90\%$ was achieved. The impact of a probability-based active learning selection for different classifiers was compared to a completely random sampling scheme based on the average of 10 independent runs, each with a different random initialization set $N_\mathrm{I}$.

*2.8.3. Features*

In this work, we only focused on the MR image foreground, because it provides already sufficient information for the classification task. The derived features were examined via a cross-correlation coefficient map to gain insight into diversity and uniqueness of the extracted features. Significance of the features, i.e. if there exists a strong relationship between the features, is shown by a two-sample t-test with derived $p$-values and a signifance level of $\alpha$. All of the 255 combinations of the 8 feature groups are investigated with optimally selected feature size $R$ after PCA in order to maximize test accuracy.

After the extraction of $F = 17386$ features from the minimal training pool $\mathcal{L}$ determined by active learning, a PCA was applied for feature reduction by linearly combining the feature space along the dominant eigenvector directions. The minimal sufficient feature space size $R$ was optimized according to the test set accuracy ACC.

*2.8.4. Classification*

Concerning the classification, the SVM classifier was compared against a DNN classifier. The parameters $C$ and $\gamma$ of the SVM were automatically optimized via a 10-fold cross-validation on the training pool $\mathcal{L}$ using a grid search in a range of $C = \{2^{-5}, 2^{-4}, \ldots, 2^{15}\}$ and $\gamma = \{2^{-15}, 2^{-13}, \ldots, 2^3\}$. The SVM was implemented using the LIBSVM library [81]. For the DNN, we used the keras library for Theano [82]. The number of layers $M = 5$ and the number of neurons in the hidden layers $N_n^{(2)} = 140, N_n^{(3)} = 120, N_n^{(4)} = 120$ of the DNN were predetermined by the Baum-Haussler rule [83] according to input size of the feature space $R$. For all experiments, the learning rate and optimizer settings were empirically determined and fixed to 0.001 with a batch size of 64 and ADAM parameters $\beta_1 = 0.9, \beta_2 = 0.99, \epsilon = 10^{-8}$ according to [71]. The dropout ratio per layer in a range of $\{0.3, 0.31, \ldots, 0.5\}$ and the regularization penalty in a range of $\{10^{-5}, 2 \cdot 10^{-5}, \ldots, 10^{-2}\}$ were optimized via a 10-fold cross-validation on the training pool $\mathcal{L}$ using a grid search.

All simulations were run on an Intel Core i7-3770 3.4 GHz with 16 GB RAM except that the DNN was run on a NVIDIA Titan X GPU.

For a varying input size $R$ (amount of principal components), the receiver operator characteristic (ROC) was plotted as true positive rate (TPR) over false positive rate (FPR) in a one-against-rest manner and the area under the curve (AUC) of the ROC was determined. The behaviour of the ACC over changing input size $R$ was also examined.

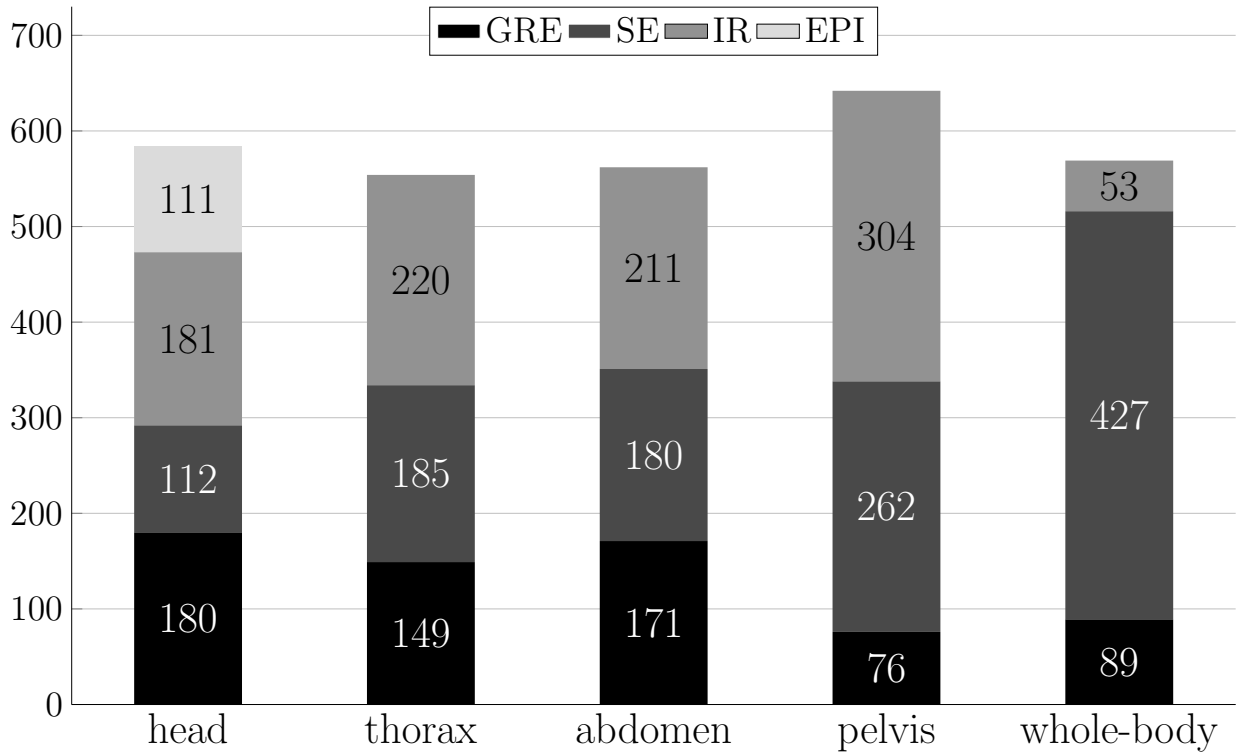

Figure 3: Image distribution of test-case database $\mathcal{X}$ consisting of 2911 datasets (250 patients, 120 female) over anatomic body regions and imaging sequences: gradient echo (GRE), spin echo (SE), inversion recovery (IR) and echo-planar imaging (EPI).

## 3. Results

In the labeling, a class label distribution of 301 labels for class 1, 621 labels for class 2, 844 labels for class 3, 837 labels for class 4 and 308 labels for class 5 was obtained. The HOs who labeled the training and test database are very consistent with a weighted Fleiss' Kappa of $\kappa = 0.93$ which corresponds to almost perfect agreement [84]. The maximal absolute deviation over all HOs and all datasets to the final decided label is just 1 class.

The overall performance of the automatic MR image quality assessment with respect to the number of features $R$ is illustrated in Fig. 4. By taking the first $R = 45$ principal components, the SVM classification reaches a maximal test accuracy of 91.3% and the DNN classification achieves an improved test accuracy of 92.5%. A slightly improved accuracy value of 93.0% can be obtained for the DNN classification by using the first $R = 47$ principal components. For a fair comparison all following values are determined for $R = 45$ principal components, if not stated otherwise. For the best classifier parametrization, the confusion matrices of the SVM and DNN are shown in Table 3. As it can be observed, each class can be well delineated from the others with only a small number of false negatives and false positives between neighbouring classes. For visual comparison, an exemplary image of the pelvis for different reconstruction parameters of proximal average with and without total variation regularization, FOCal underdetermined system solver with and without total variation regularization and of CS matching pursuit which correspond to different image quality classes is shown in Fig. 5.

The ROCs in Fig. 6 and Supplementary Fig. 1 depict the impact of the rainbow color-coded input size $R$ on the classification performance. It can be seen, that all parametrizations achieve a better performance than random guessing which corresponds to the bisecting line in the ROC, i.e. line of no-discrimination, with slightly more consistent results for the SVM. The optimal point of operation for which both classifiers achieve the maximal accuracy is highlighted. For the best parametrization an average AUC over all classes of 0.989 with an ACC of 91.3% for the SVM and of 0.994 with an ACC of 93.0% for the DNN is obtained (see also Table 3). The accuracy change for different input sizes $R$ in Fig. 4 substantiates the more consistent results of the SVM with smaller variations and higher accuracy floor for large input sizes.

Training the classifiers with the full training set of $|\mathcal{L}| = 2038$ datasets including the automatic parameter optimization takes on average $720 \pm 10$s for the SVM and $696 \pm 22$s for the DNN and excluding the parameter optimization $1 \pm 0.5$s for the SVM and $15 \pm 2$s for the DNN. Classifying a new unknown image from already extracted features can be achieved within less than a second for both classifiers.

The impact of the active learning is shown in Fig. 7. In contrast to a pure random sampling which is equivalent to normal labeling, i.e. no active learning, the probability-based selection rule helps to reduce the labeling effort. Using active learning, a test accuracy of over 90% is achieved on average for all examined parametrizations and runs with 51% less datasets for the SVM classification and with 53% less datasets for the DNN classification with very consistent results, i.e. small standard deviation over the different initialized runs. In other words for this experiment approximately 1000 datasets are sufficient for a reliable training to differentiate diagnostic usability on a 5-point Likert-scale for accelerated MR acquisitions.

The impact of various feature groups can be examined in Fig. 8. Some sets resulted in an accuracy below 50%, i.e. being no longer able to reflect and capture the underlying problem, while others reached accuracies in the range of $65-90$%. For nearly the same amount of features, some combinations yield a higher accuracy than others. Highest accuracies were obtained for a sufficiently large amount of features whereas the lowest accuracies were reached when only one feature group was selected.

The cross-correlation coefficient map of the retrieved feature vector $\tilde{\underline{x}}$ from the minimal possible training pool $|\mathcal{L}| = 1000$ determined by active learning is displayed in Fig. 9 with the intensity encoding the correlation. The null hypothesis test yields a significance of $p < 0.05$ in 63% of all features and a significance of $p < 0.01$ in 58% of all features, i.e. approximately 60% of the features are correlated with a high probability. This suggests to leave out the correlated features and investigate the impact on the classification. For a manually reduced feature set of $F = 6433$ features containing only the significant and non-redundant features, a test accuracy of 92.4% with $R = 46$ for the SVM and of 93.7% with $R = 50$ for the DNN is obtained. By cropping the feature set further down to only $F = 2871$ features containing the gradient, Gabor, run length, fractal dimension, GLCM and LBP features, a test accuracy of 91.2% with $R = 77$ for the SVM and of 92.7% with $R = 64$ for the DNN is achieved. The full feature vector $F = 17386$ can be retrieved on average within $134 \pm 4$s.

## 4. Discussion

We propose a generic and automatic MR image quality assessment framework which is able to provide feedback about the derived image quality in relation to a certain underlying question/application. Although the individual components are already known in literature, some components as the feature extraction, DNN and active learning were extended and modified to work with the provided MR image input. Moreover, the novelty of this work lies in the conjunction of the individual components to provide a reliable prediction. The reliabilty and stability is examined in detail in this work.

Table 3: Confusion matrices for the best parametrizations with (a) $R = 45$ for SVM and (b) $R = 47$ for DNN from full feature set $F = 17386$ in reduced training database $|\mathcal{L}| = 1000$.

|  | predicted class $\hat{y}$ | | | | |
|---|---|---|---|---|---|
| labeled class $y$ | 1 | 2 | 3 | 4 | 5 |
| 1 | 69 | 28 | 0 | 0 | 0 |
| 2 | 18 | 175 | 3 | 0 | 0 |
| 3 | 0 | 2 | 243 | 6 | 0 |
| 4 | 0 | 0 | 15 | 238 | 0 |
| 5 | 0 | 0 | 0 | 4 | 73 |

(a) SVM

|  | predicted class $\hat{y}$ | | | | |
|---|---|---|---|---|---|
| labeled class $y$ | 1 | 2 | 3 | 4 | 5 |
| 1 | 62 | 35 | 0 | 0 | 0 |
| 2 | 0 | 194 | 2 | 0 | 0 |
| 3 | 0 | 2 | 241 | 7 | 0 |
| 4 | 0 | 1 | 11 | 240 | 1 |
| 5 | 0 | 0 | 0 | 2 | 75 |

(b) DNN

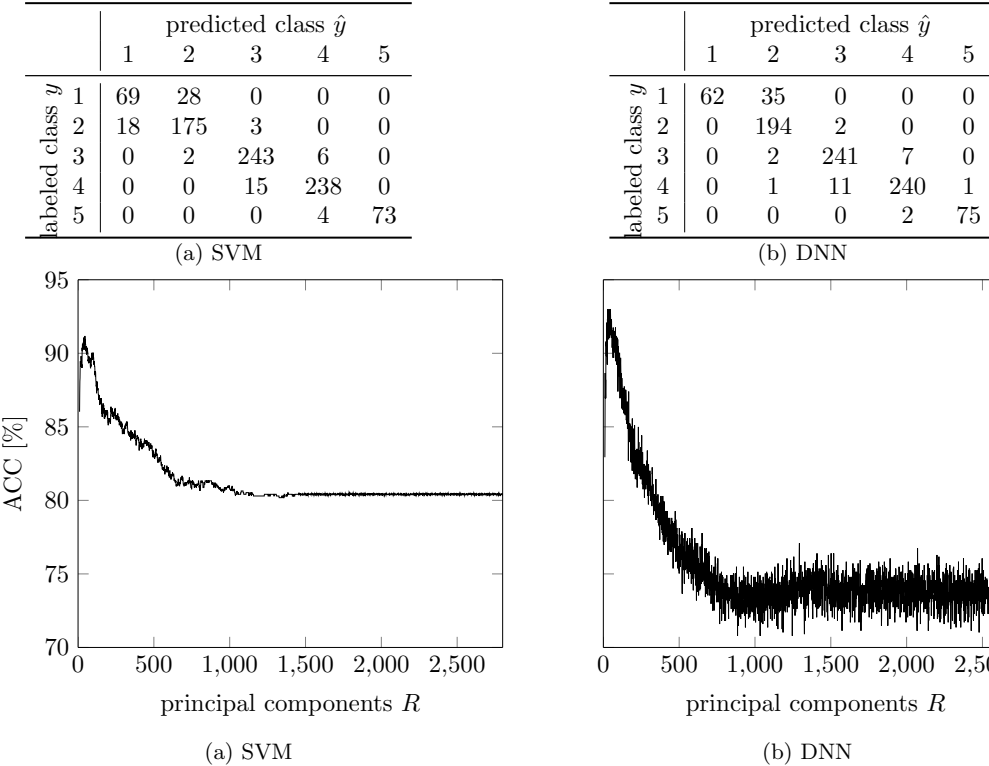

(a) SVM

(b) DNN

Figure 4: Test accuracy of (a) SVM and (b) DNN classifier for changing input feature size/amount of principal components $R$ in reduced training database $|\mathcal{L}| = 1000$.

4.1. Database and labeling

The framework is trained with labeled datasets of HOs who quantify the quality with respect to the underlying question at hand. It is attempted to keep the quantification objective, i.e. minimizing inter-reader variability, through the provided guidance in order to get a reliable label. Furthermore, intra-reader consistency is also very important which we ensure by assisting the HO in the labeling procedure, i.e. by enabling the viewing of previously labeled datasets of the same or a similar test-case. This necessity can be acknowledged if one observes the challenging task to assign a label for the datasets in Fig. 5. So a rough guidance and assistance is of great help. The developed website helps to provide a very convenient and easily accessible labeling platform which reflects in a faster and easier allocation of labels. The provided website functionalities are sufficient to determine an appropriate label.

For the labeling we have decided to account for different labels, i.e. if the HO do not agree, we calculate the median as a robust and discrete decision rule. However, the HOs were quite consistent with an average label distance of just 1 class which is also indicated by a high Fleiss' Kappa value of inter-HO agreement. A high value of $g = 4 = \#\text{HO} - 1$ was chosen to approve that all HOs except for one always coincide. In the future, we plan to experiment with a weighted mean of the individual labels based on the years of experience, but up to now we tried to keep the label decision simple in order to trust the derived labels and focus on the classification task.

For the task of perceivable image quality assessment, we have only focused on global labels, i.e. a score for the whole 3D volume, because a local artifact affects and depreciates the whole acquired volume. It shall be noted that local labels (e.g. about the actual occurrence of the artifact) provide more detailed information and can improve the classification [85, 86].

The a-priori selection of suitable datasets from the whole database for the test-case database $\mathcal{X}$ helps to improve the labeling consistency and the robustness of the classification task. Moreover, test-case unrelated datasets may mislead the classification task, because the active learning might select these datasets as new distinctive information which can result in longer runtimes of the MR image quality assessment framework till the stopping criteria is reached. Hence, a proper preselection helps also to reduce the computational complexity. In the conducted experiment we chose an approximately uniformly representation of the data amongst the most commonly used imaging sequences equally distributed for all body parts. The image distribution over the body regions and sequences results from the clinical routine protocols which are usually conducted in these anatomical regions, e.g. less EPI sequences for static anatomical imaging in the body trunk. Furthermore, for this study we just focused on static imaging in order to investigate the nature of the classification in the feature space. To analyze the discriminative potential of the framework, datasets from different imaging sequences were used instead of a specific sequence. In the future however, studies may focus on specific sequences to derive meaningful features or the inclusion of dynamic and functional imaging might bring additional interesting dimensions (temporal, parametric, ...) for exploration.

Due to varying patient compliance, some images exhibit more artifacts than others for the same sequence parametrizations. All subsampling experiments were conducted prospectively to fully appreciate the impact of aliasing and other related artifacts. The percentage splitting of the test-case database into a training and test set is based on classical empirical values. However, similar classification results with a test accuracy of above 90% can also be obtained if specific patients, e.g. the ones with the most consistent labeling or with visually low image quality due to artifacts, are moved from the training to the test set. This indicates that the current setup does not tend to an overfitting in the training and is still generic enough to capture and classify new unseen images correctly.

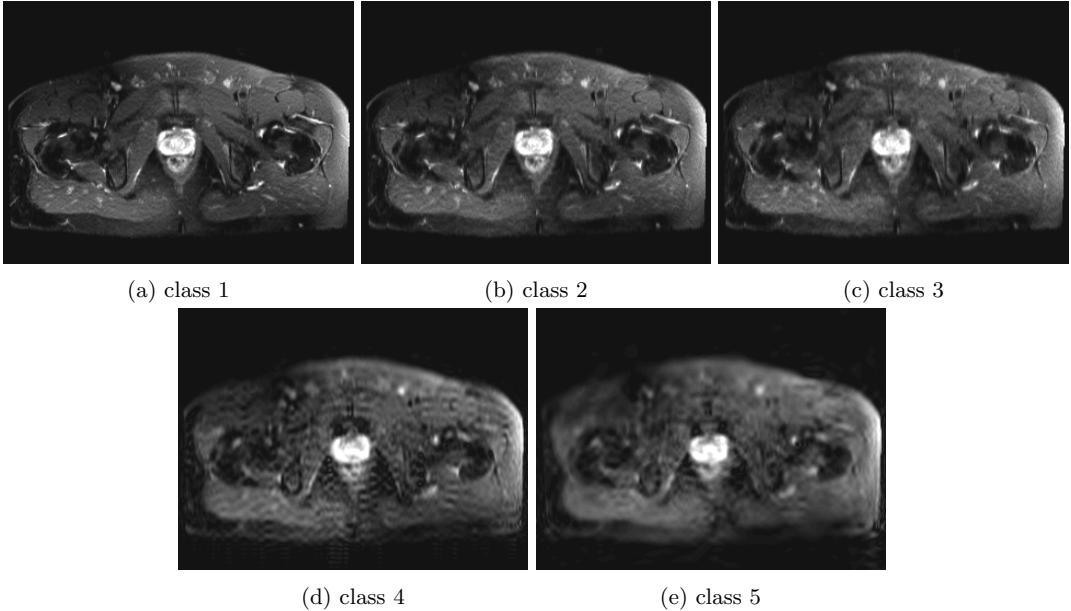

(a) class 1  (b) class 2  (c) class 3

(d) class 4  (e) class 5

Figure 5: Visual comparison of the different quality classes of a pelvic image for changing Compressed Sensing reconstruction algorithms: proximal average method without and with total variation regularization, FOCal Underdetermined System Solver with and without total variation regularization, compressive sampling matching pursuit; from class 1 (best) to 5 (worst).

4.2. Active Learning

The incorporation of active learning is motivated by the requirement to reduce the effort in the manual labeling. Active learning avoids the selection of redundant datasets to improve the classification result with contrasting and distinctive new information. Datasets are chosen based on the uncertainty of the classifier. This selection does not only improve the classification results faster, i.e. using less labeled samples, but also helps to keep the HO on track and longer motivated to proceed with the labeling process by viewing new and changing image contents. While at the moment distinctive image contents are selected based on the database composition and the level of classification uncertainty, this might be further improved in the future to constrain the selection also based on the image content. The initial training size $N_\mathrm{I}$ is a crucial factor for the active learning with respect to labeling effort and achievable improvement. A small training set can be labeled fast, but if the set is too small, the resulting accuracy is not very promising and hence the HO is queried more often compared to a larger initial training set. On the other hand, a too large initial training set reduces the benefit of active learning. As we already reported in [35], changing the initial training size does not affect the incremental accuracy improvement of the active learning. Hence, the selection of $N_\mathrm{I} = 200$ initial training samples was a tradeoff between initial labeling effort, amount of HO queries and incremental accuracy improvement. The same observations hold for the size of the query set $|\mathcal{Q}|$. If chosen too small, the HO is queried too often, and if chosen too large, the framework might not adapt to changing classification conditions fast enough, yielding a reduced incremental improvement of the accuracy. This can also be observed in Fig. 7, where for larger $|\mathcal{Q}|$ the curves tend to flatten for some update steps while the small set $|\mathcal{Q}| = 20$ has a more monotonic increase. However, all query sizes $|\mathcal{Q}|$ reach the goal of ACC $> 90\%$ at nearly the same amount of labeled datasets. Hence, changing the size $|\mathcal{Q}|$ has just little influence on the to be labeled datasets, but depending on the computation time of the framework it might have a negative side effect on the perceived performance, i.e. feedback to the website if the framework is blocked by slow classification updates. If the HO labels faster than the next query set is available, then the HO either has to wait or will label randomly picked datasets which both is not desirable. Therefore, based on an approximate computation time in the range of $40 - 60$s (with pre-derived features and restricted automatic classifier parameter tuning), a sufficiently large query set should be chosen. In our experiments, a size of $|\mathcal{Q}| = 40$ yields the best tradeoff and meets the requirements. The active learning strategy can also be employed with other query strategies such as expected model change or model error reduction, but this would require a model of the classification task and feature space which might be difficult to set up for changing test-case questions and databases and is hence more probably prone to errors. In contrast to our previous work [35], we focused on the probability-based uncertainty sampling, because it allows a direct comparison of SVM and DNN classification with comparable results to the previously reported feature-space distance-based selection. A labeling effort reduction of approximately 50% can be achieved indicating that the classifier benefits from training on non-redundant images respectively extracted features. It shall be noted that this reduction may vary for different databases, quality of HO labels and extracted features.

*4.3. Foreground/background segmentation*

We decided to use a fast and robust Chan-Vese segmentation to separate the MR image foreground from the background. Not only does this allow for a separate investigation of descriptive features for image content (foreground) and scanning conditions (background) such as scanning hardware or patient impact on scan (e.g. respiratory motion), but it also stabilizes the extracted features and makes the framework more robust. If no separation was performed, mainly regional and intensity features tend to have large variations for changing background and similar image contents which make a reliable class prediction very challenging. Furthermore, the separation also helps to reduce redundant background information if the foreground is observed, yielding a more stable feature set. Additionally the computation can be carried out faster for a reduced image size. Of course the background carries also some information about artifact burden, but we found that the foreground provided enough information by e.g. aliased content to differentiate the quality classes. The addition of background features just slightly improved the classification accuracy by $< 1\%$, but demanded more careful feature selection which will be examined more thoroughly in future studies. Therefore, in this experiment we only considered the foreground as it already provided sufficient information to differentiate the image-series.

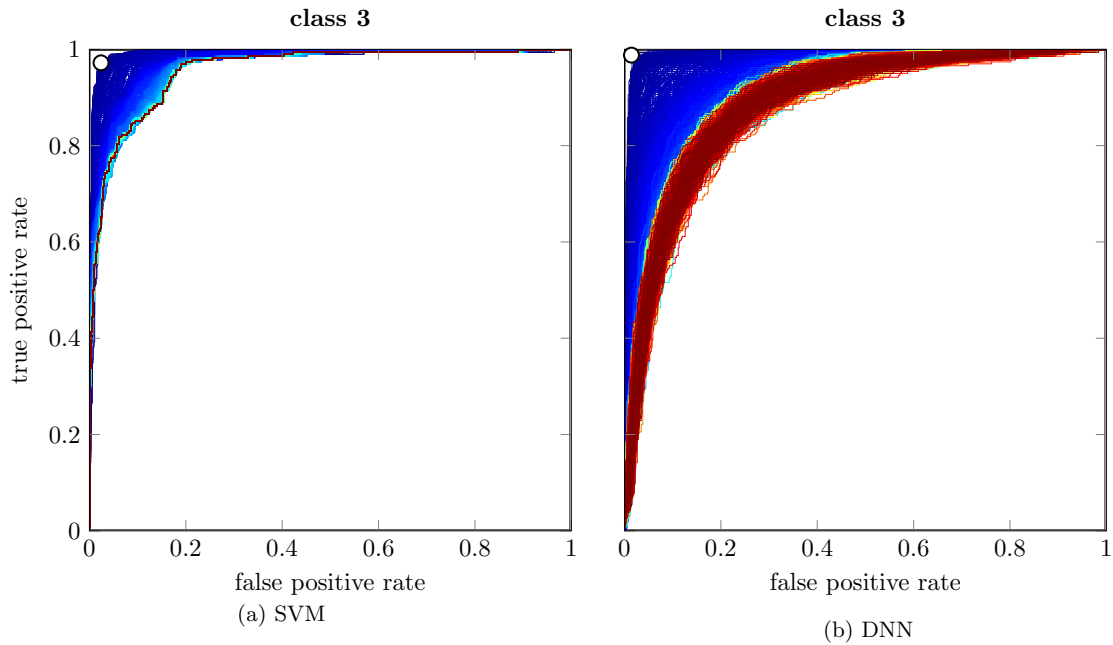

Figure 6: Receiver operating characteristic for class 3 in (a) SVM and (b) DNN classifier in a one-against-rest manner for increasing input feature size $R$ (rainbow color-coded from blue to red) with the marked optimal operating point in the sense of highest test accuracy. The area under the curve (AUC) is in class 3 for the SVM: $AUC_3 = 0.994$ and for the DNN: $AUC_3 = 0.995$. The receiver operating characteristics for all classes are shown in Supplementary Figure 1.

4.4. Features

Features are extracted from the image foreground. No preprocessing such as resampling, scaling or denoising is performed in order not to interfere with data fidelity. In order to provide sufficient diversity and being able to capture all subtle changes, a large amount of features is extracted at first. Since our MR image quality assessment framework aims to provide a generic framework to classify any kind of input images with changing questions/applications in a reliable and robust way, a large feature database is constructed. In this experiment we were interested in the overall performance of this large feature database on the minimal necessary training pool, i.e. if any of the features provide more or diverse information than others. We therefore tried to group the features as shown in Fig. 2 and Table 1 based on their extracted image properties to pave the way for a manual feature tuning and selection. Automatic selection techniques as for instance sequential floating forward selection (SFFS) are not applicable due to its high combinatorical complexity. As illustrated in the correlation coefficient map in Fig. 9, some features show a high correlation, e.g. those in the region group namely between the EBR and LBP features. This indicates that not all features are needed to retrieve a reliable classification, because they carry redundant information. This correlation occurs, because a lot of features concentrate solely on the image intensity, but lack the incorporation of any neighborhood, regional or descriptive information which might be superior for the differentiation. Thus, from the complete feature set, we selected several smaller subsets and investigated their performance. On average, the more features being considered, the better the performance. However, certain smaller-sized combinations yielded also very high accuracies. This indicates the room for improvement because of redundant features. Overall, the intensity, geometrical and regional feature groups contribute the most to the framework's performance. Here we consider two prominent subsets: one containing only the uncorrelated features with a significance of $p > 0.05$ from the complete set yielding $F = 6433$ (set A) and one containing specifically selected features which show a low cross-correlation, namely gradient (intensity), Gabor (transformation), run length (geometrical), fractal dimension (geometrical), GLCM (geometrical) and LBP (regional) features yielding $F = 2871$ (set B). For the subsequent PCA feature reduction, the best resulting feature size $R$ is determined with automatic classifier parameter optimization. While the full feature set $F = 17386$ achieves an accuracy for the SVM/DNN classification of 91.4%/93.0%, the feature set A with $F = 6433$ achieves an accuracy of 92.4%/93.7% and the feature set B with $F = 2871$ achieves an accuracy of 91.2%/92.7%, respectively. As can be seen, reducing the feature set to the most significant features (set A) helps to increase the accuracy. However, a further reduction to only non-redundant features (set B) results in a slightly reduced accuracy compared to the full set. So resolving the dependency between the features and focusing on only significant and informative ones improve the test accuracy. Nevertheless, a slight feature overlap, i.e. feature correlation, can also be beneficially exploited by the classifier as observed by going from set A to set B. If the feature set composition is not chosen carefully, the test accuracy may even drop below 50%, i.e. worse than random guessing. So the experiments indicate that a sufficiently large feature set is necessary to achieve a reliable and robust quality estimation. The specificity and robustness of the feature set need to be validated further in the future by adding patients from other cohorts and/or applications to the test-case database. At the moment, the highest computational burden lies on the feature extraction step, but so far no computational optimization was carried out, leaving room for future improvements. Furthermore, some features can be improved by the extension to 3D and can be tuned to match better the specific questions or can be replaced by features derived from feature learning.

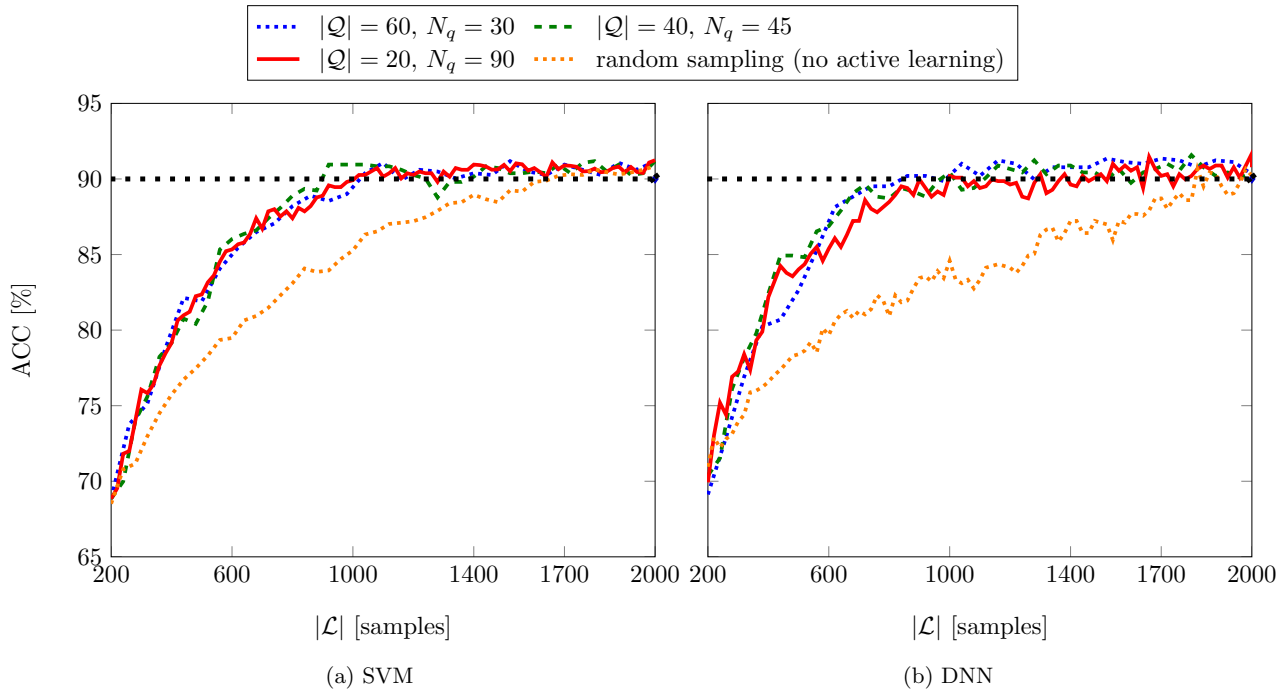

Figure 7: Active learning impact on test accuracy over $|\mathcal{L}|$ samples in the training pool and for $|\mathcal{Q}|$ samples per query in $N_q$ queries with an initial training size of $N_{\mathrm{I}} = 200$ and for the first 45 principal components. Curves represent average values over 10 different initialized runs. The (min/mean/max) standard deviation is for the (a) SVM classification: ⋯⋯ (0/0.78/2.72), - - - (0.13/0.85/2.54), —— (0/0.67/2.54), ⋯⋯ (0/0.81/2.62) and for the (b) DNN classification: ⋯⋯ (0/0.89/2.61), - - - (0/0.86/2.81), —— (0/0.95/2.94), ⋯⋯ (0/0.82/2.71).

For the feature reduction, a PCA is employed as a robust and fast method. The PCA has the advantage that the feature information from the complete feature space is still utilized by the linear feature combination along the dominant eigenvectors, i.e. combining the dominant energy components instead of throwing them away completely. Before the PCA, a zero-mean unit-variance scaling is applied to scale all individual features. Fig. 4 demonstrates that a high accuracy is obtained if the input size is $R \leq 100$. Thus a remarkable reduction of the feature dimensionality can be achieved which is also advantageous for the classification, because it reduces the computational complexity tremendously. This large reduction factor can already be appreciated by the strong cross-correlation between some of the features. The experiments also demonstrated that for decreasing feature size $F$, the classifier input size $R$ after feature reduction increases a little bit, because feature diversity gets lost and hence more features need to be combined to achieve the same classification performance.

4.5. Classification

In the conducted experiment, two widely used classifiers SVM and DNN are chosen because of their strong generalization ability, good capabilities to differentiate non-convex class clusters, fast and efficient algorithms for large-scale inputs and their possibility to deal with the multi-class case. The classifiers are trained on manually labeled images which are rated by HOs and do not need any reference image.

Both classifiers show a similar performance if parametrized suitably, with a small advantage for the DNN. However, the DNN also requires a larger training set for solid learning which at the moment could just be fulfilled if the degrees of freedom (number of hidden layers and neurons per layer) are limited. This means that the full potential of the DNN might not yet have been fully observed. Furthermore, the computational complexity of the DNN is much larger than the SVM and requires the computations to be performed on a GPU. The computation time in the training of the classifier depends on the allowed parameter ranges for the hyperparameter optimization and the input size $R$. Since the feature reduction step suggests to use an input size of $R \leq 100$, the main computational burden comes from the hyperparameter optimization. In our experiments these parameter ranges are set quite wide in order to guarantee a good performance of the classifier and are not yet optimized for computational speed. Hence by either reducing the parameter ranges or by fixing these parameters, one can reduce the computation time in the training further. After the training of the model, both classifier are very fast in predicting a new label.

The chosen multi-class case allows a finer distinction between different quality levels than just a binary decision as it is usually performed in other works. The quality classes are not limited to five, but have been chosen in this experiment in similarity to a Likert-scale and in order to provide enough variability for the quality classes with limited labeling uncertainty. This labeling uncertainty respectively its impact can already be appreciated if one takes a look at the confusion matrices in Table 3. While a good distinction between far away quality classes can be achieved, e.g. between class 2 and 5, the distinction between neighboring quality classes is always very challenging and there are some remaining false positives and negatives. Furthermore, the high quality classes 1 and 2 are hard to differentiate which is also very challenging for an HO as can be observed in Fig. 5. Hence this is also hard to distinguish by a classifier. To compensate for the imbalanced training set sizes per class and to minimize the sensitivity of the classifier for this issue, a one-vs-one approach was chosen for the SVM. It shall be noted that it is also possible to calculate a continuous-valued image score by using the a posteriori probability for a further fine-grained distinction to maintain more information about the quality levels.

The ROC analysis gives insight on the parametrization impact on the classification task for each class against the remaining ones. Only the ROCs for changing input size $R$ are shown in Fig. 6 and Supplementary Fig. 1 to enable a comparison of its dependency on the classifier. The automatic hyperparameter optimization by cross-validation and grid search helps to tune the parameters accordingly. In close proximity to the optimal hyperparameter values, the overall accuracy changes only by $\sim 5\%$, i.e. the estimation is quite stable. Further away from the optimal values, the accuracy may drop to 30%, i.e. close to random guessing. Therefore, the automatic hyperparameter optimization is a crucial step to guarantee convergence towards a reliable estimation. For the remaining parameters we observed a rather stable trend and hence excluded them from the automatic optimization. These parameters are empirically chosen lenient to satisfy the overall classifier input, i.e. database distribution and extracted features.

For all classes a nearly optimal ROC with a high AUC is obtained. Comparing the ROCs of SVM and DNN, the SVM shows a more consistent performance while the DNN shows an overall better performance for changing input size. The performance of both classifiers decreases if $R$ is too small, because the classifiers lose the ability to separate the classes in a low-dimensional feature space. If $R$ is too large, the classifiers run into a curse of dimensionality with the accuracy dropping down to a constant accuracy floor. The optimal input size $R$ is hence reached according to Fig. 4 in a range of $15 \leq R \leq 100$. For a fair comparison the number of layers and neurons in the DNN is kept fix for increasing input size $R$. An increase of the layers and neurons may improve the accuracy again, but may also lead to overfitting and should hence be avoided.

Overall the DNN slightly outperforms the SVM. In the future, the manual feature extraction shall be replaced by an automatic feature learning which can be easily incorporated into the DNN architecture.

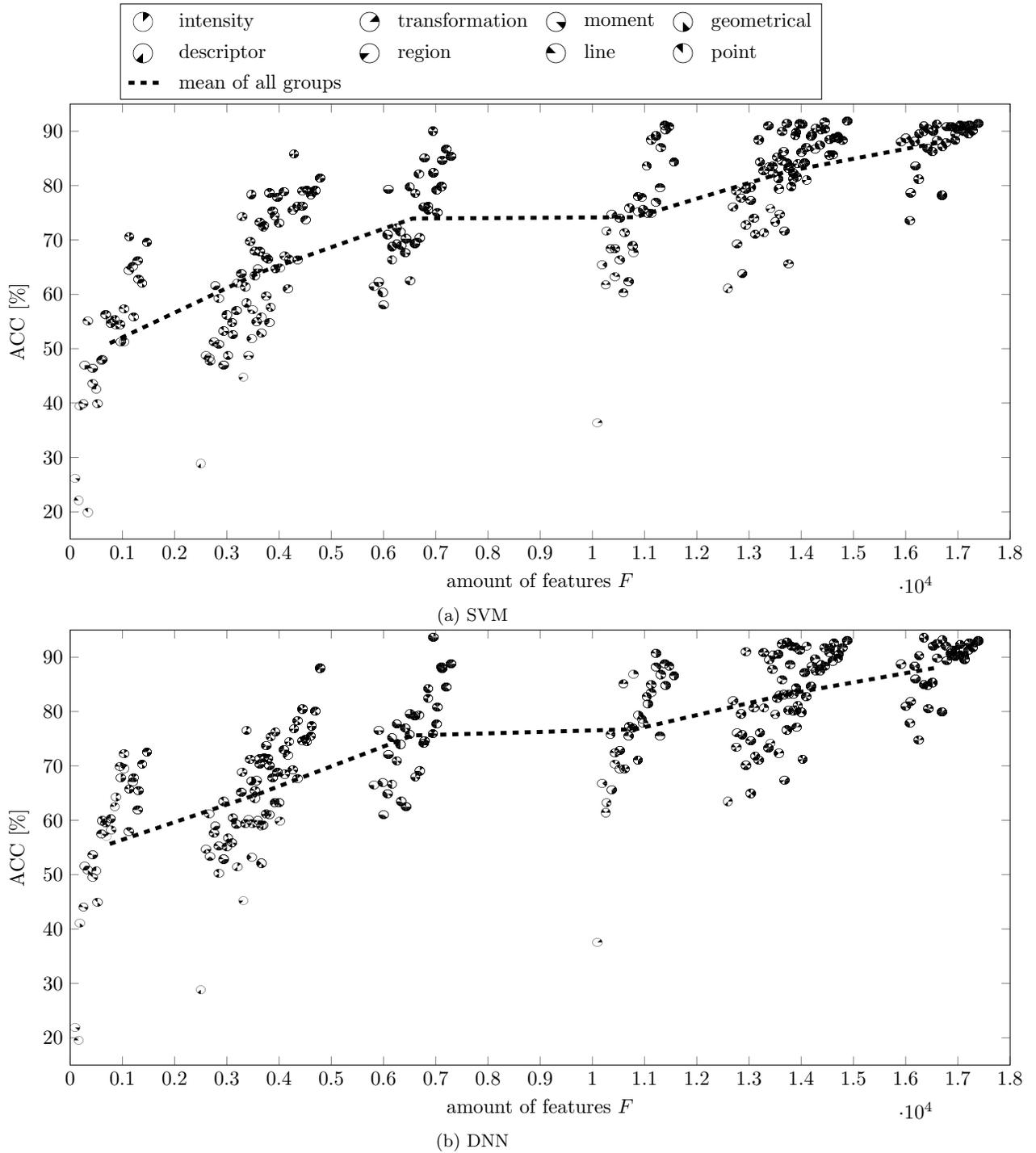

Figure 8: Comparison of different feature group combinations in terms of test accuracy ACC with (a) SVM and (b) DNN classification. The average test accuracy behaviour for various amount of features $F$ is also shown.

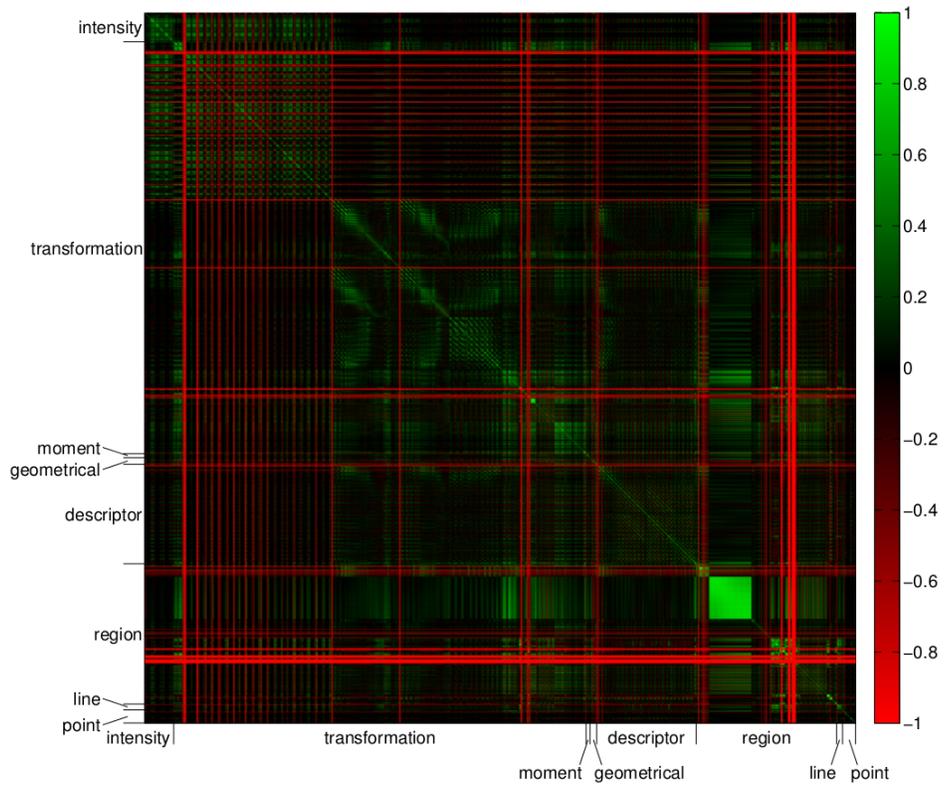

(a) cross-correlation coefficient

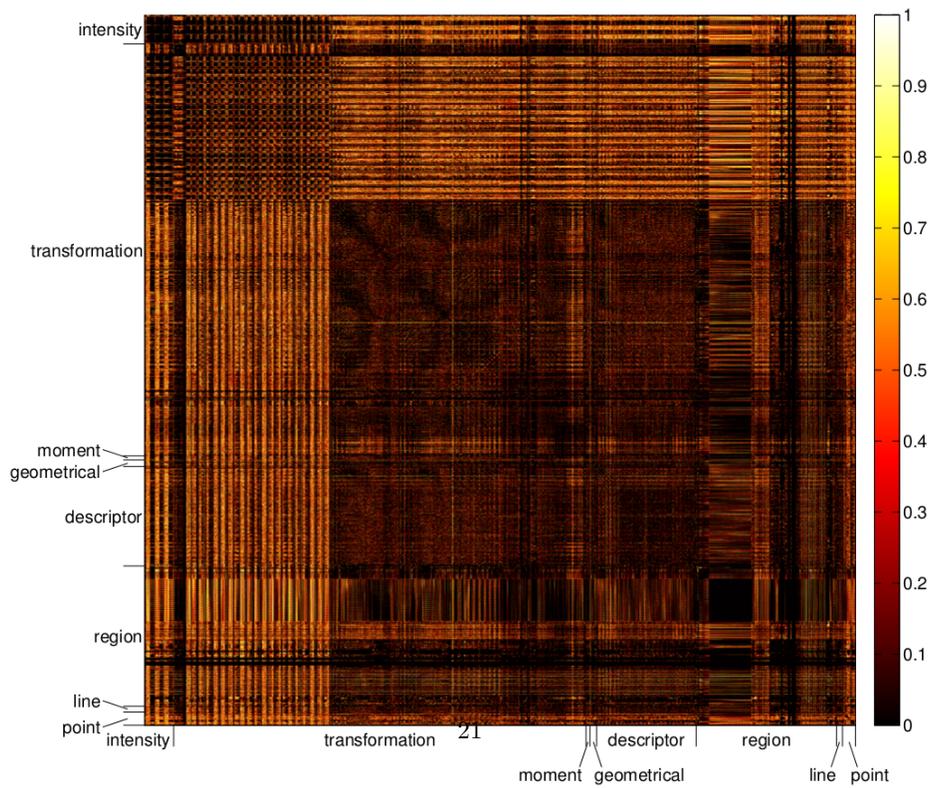

(b) p-value

Figure 9: Extracted image features $F = 17386$ from reduced training database $|\mathcal{L}| = 1000$ as (a) cross-correlation coefficient map and (b) p-values.

## 5. Conclusions

We propose an automatic reference-free MR image quality assessment framework which is able to mimic the perception of an HO and can predict MR image scores with respect to an underlying question/application. The framework was tested in a cohort of 250 patients with varying scanning and reconstruction conditions including the associated artifacts to perform a retrospective quality control of the obtained MR images. This is especially of interest for large cohort studies in order to guarantee successful post-processing.

The framework is trained on HO-derived labels. The HOs labeled the training images with respect to diagnostic usability on a 5-point Likert-scale. Since labeling can be a very time-consuming and cost-intensive task, the process was shortened by the incorporation of active learning and further streamlined by an easy-to-use labeling website. This reduced the amount of required training data by 53%. Different features were provided and tested to investigate the optimal set for the task of predicting the images' diagnostic usability. The MR image quality assessment framework achieves a high accuracy of 93.7% for the best parametrization with a DNN architecture indicating that the derived image feature set together with the classifier is able to capture quality changes for different imaging sequences, contrasts, body regions and artificial burden.

## Acknowledgements

The authors would like to thank Christina Schraml, Cornelia Brendle and Ferdinand Seith for their effort in the labeling step and for helpful discussions. Furthermore, we thank Carsten Gröper and Gerd Zeger for assisting in data acquisition and Brigitte Gückel for study coordination. We gratefully acknowledge the support of NVIDIA Corporation with the donation of a GPU used for this research.

**Supplementary figures**

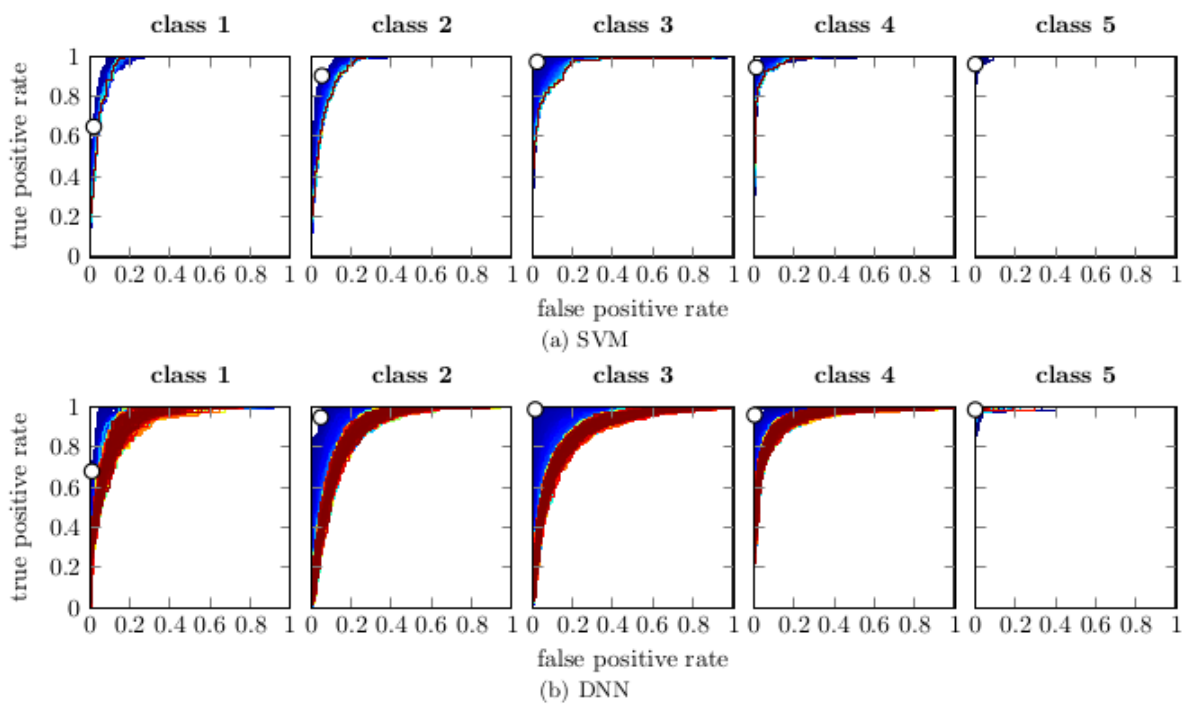

Supplementary Figure 1: Receiver operating characteristic of (a) SVM and (b) DNN classifier and for each class in a one-against-rest manner for increasing input feature size $R$ (rainbow color-coded from blue to red) with the marked optimal operating point in the sense of highest test accuracy. The area under the curve (AUC) for the optimal operating point of classes 1 to 5 in the SVM is $AUC_1 = 0.976, AUC_2 = 0.980, AUC_3 = 0.994, AUC_4 = 0.995, AUC_5 = 0.999$ and in the DNN is $AUC_1 = 0.989, AUC_2 = 0.993, AUC_3 = 0.995, AUC_4 = 0.993, AUC_5 = 0.999$.